\documentclass[lettersize,journal]{IEEEtran}
\usepackage{amsmath,amsfonts}
\usepackage{algorithmic}
\usepackage{algorithm}
\usepackage{array}
\usepackage[caption=false,font=normalsize,labelfont=sf,textfont=sf]{subfig}
\usepackage{textcomp}
\usepackage{stfloats}
\usepackage{url}
\usepackage{verbatim}
\usepackage{graphicx}
\usepackage{cite}

\usepackage{amssymb}
\usepackage{booktabs}
\usepackage{epsfig}
\usepackage{multirow}

\newcommand{\vect}[1]{\mathbf{#1}}
\newcommand{\matr}[1]{\mathbf{#1}}

\newcommand{\set}[1]{\mathcal{#1}}

\begin{document}

\title{Revisiting Pretraining for Semi-Supervised Learning in the Low-Label Regime}

\author{Xun Xu,
        Jingyi Liao,
        Lile Cai,
Manh Cuong Nguyen,
Kangkang Lu,
Wanyue Zhang,
Yasin Yazici,
Chuan Sheng Foo
\thanks{X. Xu, J. Liao, C. Nguyen, K. Lu, Y. Yazici and C.S. Foo are with the Institute for Infocomm Research (I2R), A-STAR, Singapore. W. Zhang is with Max Planck Institute for Informatics.}
}

\markboth{Journal of \LaTeX\ Class Files,~Vol.~14, No.~8, August~2021}%
{Shell \MakeLowercase{\textit{et al.}}: A Sample Article Using IEEEtran.cls for IEEE Journals}


\maketitle

\begin{abstract}
 Semi-supervised learning (SSL) addresses the lack of labeled data by exploiting large unlabeled data through pseudolabeling. However, in the extremely low-label regime, pseudo labels could be incorrect, a.k.a. the confirmation bias, and the pseudo labels will in turn harm the network training. Recent studies combined finetuning (FT) from pretrained weights with SSL to mitigate the challenges and claimed superior results in the low-label regime. In this work, we first show that the better pretrained weights brought in by FT account for the state-of-the-art performance, and importantly that they are universally helpful to off-the-shelf semi-supervised learners. We further argue that direct finetuning from pretrained weights is suboptimal due to covariate shift and propose a contrastive target pretraining step to adapt model weights towards target dataset. We carried out extensive experiments on both classification and segmentation tasks by doing target pretraining then followed by semi-supervised finetuning. The promising results validate the efficacy of target pretraining for SSL, in particular in the low-label regime.
\end{abstract}


\section{Introduction}
\label{sec:intro}

Deep neural networks trained on large collections of labeled data have achieved unprecedented performance on many computer vision tasks. Unfortunately, acquiring these large amounts of labeled data is expensive and laborious. One effective strategy to mitigate this dependence on labels is exploiting unlabeled data for training, as done in semi-supervised learning (SSL).

Semi-supervised learning tackles the challenge of limited
labeled data by exploiting a large amount of unlabeled
data. Recent consistency-based SSL methods obtain pseudo-labels for unlabeled data through ensembles of predictions~\cite{samuli2017temporal}, network parameters~\cite{tarvainen2017mean} or weaker augmentations~\cite{sohn2020fixmatch}. A student network is further trained on both labeled data and unlabeled data with pseudo-labels. Since these methods are directly trained on a specific task (dataset), we refer to this overall approach as task-specific SSL \cite{chen2020simple}.
Task-specific SSL has demonstrated very competitive performance on simpler datasets like CIFAR-10~\cite{krizhevsky2009learning}. However, the performance of task-specific SSL in more challenging settings, for instance, fine-grained classification~\cite{wang2021self}, the extremely low-label regime~\cite{wang2021self} and segmentation~\cite{french2019semi}, is still far from satisfactory. This is partly because when few labels are available, there is simply not enough supervision to generate accurate pseudo-labels, thus misleading the training of student network~\cite{arazo2020pseudo}. 
 
 \begin{figure}
     \centering
     \includegraphics[width=1\linewidth]{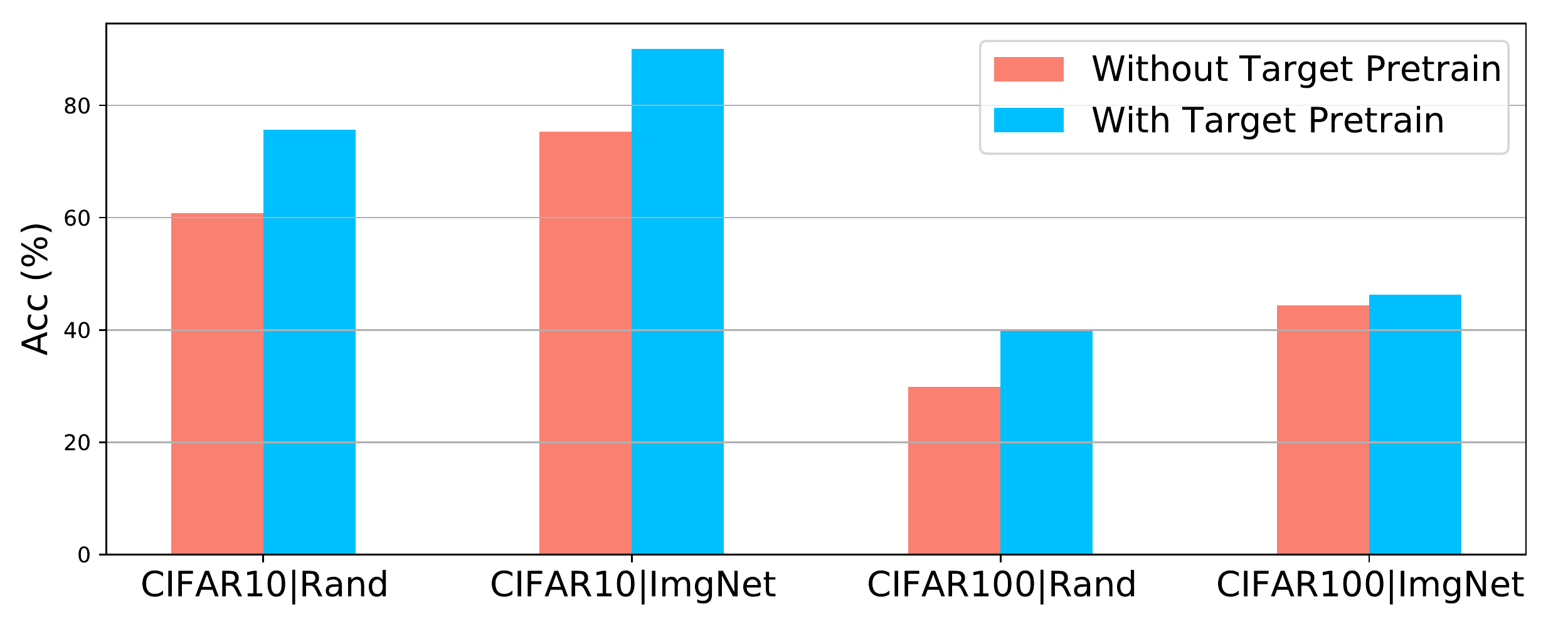}
     \caption{Illustration of FixMatch finetuning from different initial weights on CIFAR-10 (40 labeled data) and CIFAR-100 (400 labeled data). Target pretraining on target datasets (CIFAR-10 \& CIFAR-100) improves semi-supervised learning performance respectively. }
     \label{fig:thumbail}
 \end{figure}

In contrast to task-specific SSL, the finetuning (FT) approach involves pretraining a backbone network on a separate, large dataset in either supervised~\cite{he2019rethinking} or unsupervised~\cite{chen2020simple,grill2020bootstrap,he2020momentum,henaff2020data} manner, and then finetuning the network on the potentially smaller target dataset. 
For example, it is common practice in computer vision to start with a model pretrained on ImageNet~\cite{russakovsky2015imagenet}, and then finetune it on the downstream task of interest.
The FT paradigm has been successfully used across wide range of computer vision tasks, including classification~\cite{shao2014transfer}, segmentation~\cite{chen2018encoder}, detection~\cite{ren2015faster} etc.

\begin{figure*}[htb]
    \centering
    \includegraphics[width=0.9\linewidth]{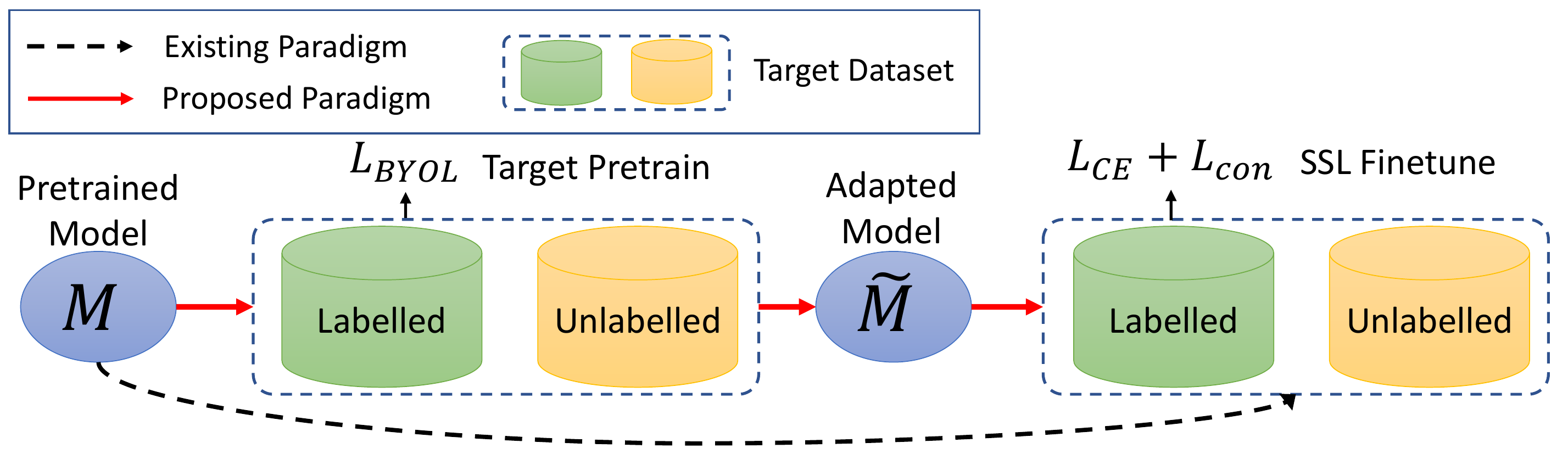}
    \caption{The pipeline for contrastive target pretraining for better initial weights. Existing works often use pretrained model weights to directly initialize semi-supervised finetuning on target dataset. In contrast, we introduce an additional adaptation step by doing contrastive target pretraining on the target dataset (both labeled and unlabeled) before semi-supervised finetuning.}
    \label{fig:intro}
\end{figure*}

The advantages of SSL and FT are often orthogonal as the former exploits unlabeled data to generate pseudo labeled data while the latter provides better initial model parameters.
Therefore, synergistically combining FT with SSL, termed as finetuning + SSL, to enjoy the benefits of both strategies, has been explored by recent works. Among these attempts, SimCLR~v2~\cite{chen2020big} starts with ImageNet pretrained weights and then employs a supervised loss to finetune on a small labeled target dataset followed by distillation on unlabeled data. This initial attempt of finetuning + SSL in a non-trivial way demonstrated superior performance, compared with task-specific SSL methods, mainly in the low-label regime with very large network capacity. Along this direction, Self-Tuning~\cite{wang2021self} further proposed to combine cross-entropy loss with a multi-positive pair contrastive loss during finetuning. It was claimed that by incorporating this additional contrastive loss with pseudo-labels, confirmation bias~\cite{arazo2020pseudo} and model shift~\cite{you2020co} issues can be mitigated, and Self-Tuning is reported to outperform state-of-the-art semi-supervised learners.

The reported large performance improvement of finetuning + SSL~\cite{chen2020big,wang2021self} over task-specific SSL~\cite{tarvainen2017mean,miyato2018virtual,xie2019unsupervised,sohn2020fixmatch}, inspired us to explore the following questions: 1) What is the key ingredient that enables the success of finetuning + SSL? 2) Does it generalize to task-specific SSL? 3) Is task-specific SSL still the most effective SSL approach? 
To answer these questions, we first make a critical observation that one major issue with the comparison of finetuning + SSL and task-specific SSL is the initialization of model weights. In particular, SimCLRv2~\cite{chen2020big} and Self-Tuning~\cite{wang2021self} employed model weights pretrained on ImageNet while task-specific SSL methods are often trained with randomly initialized weights. 
Therefore, we hypothesize that the better \textit{initial weights} provided by pretraining on the separate large dataset is largely responsible for the superior performance reported by finetuning + SSL. It is thus reasonable to believe, if the same initial weights are available to task-specific SSL methods, a similar performance boost could be expected.
To test this hypothesis, we carry out empirical studies by comparing state-of-the-art task-specific SSL methods both with random initial weights and pretrained weights. As shown in Fig.~\ref{fig:thumbail}, with FixMatch we observe a clear performance gap between random initialized weights ``CIFAR-10/CIFAR-100|Rand'' and ImageNet pretrained weights ``CIFAR-10/CIFAR-100|ImgNet''. These observations suggest that the benefits of good initial weights also apply to task-specific SSL methods. With further empirical results in Sect.~\ref{sect:classificationtask}, we reveal that under fair comparison, task-specific SSL finetuning still outperforms state-of-the-art finetuning based methods by a large margin.

Based on the insights above, we further notice that pretraining is often carried out on a large pretrain dataset, e.g. ImageNet, while SSL finetuning is often on a smaller target dataset, e.g. CIFAR-10, CIFAR-100, PascalVOC. 
There is an inevitable distribution mismatch between pretrain and target dataset, commonly referred to as covariate shift~\cite{wang2018deep}. 
The backbone network optimized for pretrain data distribution is subject to covariate shift and therefore will not be optimal for direct reuse on finetuning tasks. To bridge the data distribution gap, we introduce an additional target pretraining step before semi-supervised finetuning. Target pretraining is realized by doing contrastive training on both labeled and unlabeled data in the target dataset. With this additional step, model parameters will adapt to target data distribution and eventually improve finetuning performance. In particular, we find this target pretraining step is most effective when labeled data is sparse, reflecting that confirmation bias can be mitigated by a better initial weights.

We validate the effectiveness of target pretraining on both semi-supervised image classification and segmentation tasks. As augmentation is particularly important for segmentation tasks, we further introduce a differentiable geometric augmentation module to enhance the representation learning ability for segmentation tasks. 
We summarize the contributions of this work as follows:
\begin{itemize}
    \item We reveal that good initial weights through pretraining on large dataset is universally helpful to both finetuning based and task-specific semi-supervised learning approaches.
    \item  Realizing the data distribution shift between pretrain and target datasets, we propose a target pretraining step to adapt pretrained weights to target dataset before semi-supervised finetuning.
    \item We demonstrate on both classification and semantic segmentation tasls that the proposed target pretraining strategy can substantially improve finetuning performance at low-label regime.
\end{itemize}

\section{Related Work}

\subsection{Semi-Supervised Learning}

The success of many computer vision tasks depends on training deep neural networks with large amount of labeled data which is expensive to acquire. Semi-supervised learning (SSL) aims to address learning with small labeled data and large unlabeled data~\cite{samuli2017temporal,tarvainen2017mean,berthelot2019mixmatch,xie2019unsupervised,berthelot2019remixmatch,sohn2020fixmatch}. 

\noindent\textbf{Task-Specific SSL} is a major SSL paradigm which mainly produces pseudo-labels for unlabeled data to utilize them in training. In \cite{samuli2017temporal} pseudo-labels are produced as moving average of network predictions. To further exploit the ensemble of networks for stable and efficient pseudo-label prediction, Mean Teacher~\cite{tarvainen2017mean} proposed to utilize moving average of network parameters as teacher network which supervises a student network. More recently, data augmentation has been increasingly investigated under SSL. MixMatch~\cite{berthelot2019mixmatch} mixes up multiple images and their labels for stronger regularization for SSL. UDA~\cite{xie2019unsupervised} argues that advanced data augmentation methods play a crucial role in semi-supervised learning. More recently, FixMatch~\cite{sohn2020fixmatch} achieved significantly higher performance by simultaneously exploiting strong and weak augmentations for consistency matching. Other than classification task, SSL has been adapted to semantic segmentation tasks~\cite{french2019semi,olsson2021classmix,alonso2021semi} mainly adopting a Mean Teacher like architecture. Specialised data augmentation strategy is mainly accountable for the improvement in segmentation tasks.

\noindent\textbf{Finetuning + SSL} is an emerging approach towards SSL~\cite{chen2020big,wang2021self,su2021realistic} by combining transfer learning with semi-supervised finetuning. An initial attempt~\cite{chen2020big} in this direction pretrains the backbone network on ImageNet~\cite{krizhevsky2009learning} via contrastive learning~\cite{chen2020simple}. The network is further finetuned on limited labeled data, and then distilled on unlabeled data. Empirical study suggests the superiority of pretraining plus finetuning over task-specific SSL with small labeled data and large network capacity. Self-Tuning~\cite{wang2021self} further extended the framework by using ImageNet supervisely pretrained weights and finetune on target datasets by combining cross-entropy loss with contrastive loss. A significant improvement over task-specific SSL is observed on fine-grained classification tasks. 
In this work, observing the performance gap between finetuning + SSL and task-specific SSL, we suspect the initial weights are accountable and further reveal that improving initial weights by target pretraining are universally helpful to both paradigms.


\subsection{Domain Adaptation}

Transfer learning aims to reuse the data or model from source task/domain to target task/domain~\cite{pan2009survey}. One of the key challenges in transfer learning is the domain shift between source and target datasets and domain adaptation aims to tackle this issue. When both source and target data are available during model training, UDA~\cite{ganin2015unsupervised} introduced a discriminator to align the distribution between source and target domains. Nevertheless, the assumption of access to target domain data during model training restricts the application and recently source-free domain adaptation, a.k.a. test time training, emerges as a solution which no longer requires the access to labeled data during adaptation~\cite{kundu2020universal,sun2020test,liu2021ttt++} at the cost of modifying the learning objective during training. To avoid altering training objective, which could be very expensive for large-scale pretraining tasks, Tent~\cite{wang2020tent} adopted a self-training like approach to adapt only the affine transformation parameters for batchnorm layers by minimizing an entropy loss. Despite requiring no changes to training objective, Tent 
requires the prediction task to be consistent between source and target domains. In contrast to source-free domain adaptation and test time training, our target pretrain does not modify training objective in the source domain nor assume the prediction task to be identical between the source and target domains.
A recent related work addressed finetuning on target domain in an unsupervised manner~\cite{li2021unsupervised}. To avoid large deviations from pretrained model parameters they simultaneously finetune on both target and source domain data. Therefore, the adaptation still requires access to source domain data.

\subsection{Contrastive Learning}

 Constrative learning~\cite{chen2020simple,grill2020bootstrap,he2020momentum,henaff2020data} is particularly effective in pretraining backbone network on large unlabeled dataset by enforcing the image-wise features subject to two augmentation to be consistent. As such the network is able to attend to the discriminative areas for feature extraction. To focus on dense prediction downstream tasks, e.g. segmentation, detection, etc., contrasting pixel-wise features was proposed~\cite{xie2021propagate,wang2021dense} to train the backbone network to generate local features invariant to augmentations. Contrastive learning has been demonstrated to be friendly to data-efficient learning~\cite{henaff2020data} as the feature representation becomes more amenable for training classifiers after pretraining. However, using contrastive learning to adapt pretrained weights to heterogeneous target dataset is yet explored and in this work we employ contrastive learning as a target pretraining approach and demonstrate effectiveness when label-data on target dataset is sparse.
 

\section{Methodology}

We first provide a overview of the proposed target pretraining strategy in Fig.~\ref{fig:intro}. An initial model weights $\set{M}$ pretrained on large dataset, e.g. ImageNet, is assumed to be available. In addition, we have a target dataset consisting of labeled training samples $\set{D}_{L}=\{\matr{X}_i,\matr{Y}_i\}_{i=1\cdots N_{L}}$ and unlabeled samples $\set{D}_{U}=\{\matr{X}_j,\matr{Y}_j\}_{j=1\cdots N_{U}}$. A key step is to adapt pretrained model weights by doing target pretraining on the target dataset. In the rest of this section, we first briefly review the contrastive learning approach. 
To further improve contrastive target pretraining for dense prediction tasks, we include a differentiable affine transformation~\cite{jaderberg2015spatial} to improve the contrastive learning efficacy.


\subsection{Preliminary on Contrastive Learning}

We use $\matr{Z}=f(\matr{X};\Phi)$ to denote a backbone network, parameterized by $\Phi$, which takes input image $\matr{X}$ and produces a feature embedding $\matr{Z}$. We further denote a projection head, parameterized by $\Psi$, as $\hat{\matr{Y}}=g(\matr{Z};\Psi)$ which outputs a lower dimension embedding. Contrastive learning is formulated as minimizing a contrastive loss defined over two augmentations of the same input image. Formally, the InfoNCE loss~\cite{oord2018representation} is adopted as contrastive loss as in Eq.~(\ref{eq:InfoNCE}).
\begin{equation}\label{eq:InfoNCE}
    \mathcal{L}_{InfoNCE}=-\log \frac{\exp(s_{i,j}/\tau)}{\sum_{k\neq i}\exp(s_{i,k}/\tau)}
\end{equation}
In this formulation, the positive pair $(i,j)$ is two different augmentations of the same input image, and negative pairs $(i,k)$ are constructed as different images and $s_{i,j}$ defines a cosine similarity between feature encodings of pair $(i,j)$. It is identified in \cite{chen2020simple} that a large batchsize is essential to  contrastive learning performance which is computationally expensive.
Instead of maintaining both positive and negative pairs, BYOL~\cite{grill2020bootstrap} relies on similarity of features between a target network and an online network with target network being a parameter moving average of online network; the output of target network is used as a target for the online network to match against. Because of the slowly updating target network, trivial solution is avoided. BYOL optimizes the cosine distance, as in Eq.~\ref{eq:BYOL}, between the representation subject to two augmentations as below, where $\hat{\matr{Y}}_1=g(f(t_1(\matr{X})))$ and $\hat{\matr{Y}}_2=g(\hat{f}(t_2(\matr{X})))$ are respectively the features after projection head with two augmentations $t_1(\cdot)$ and $t_2(\cdot)$ and $q(\cdot)$ is a predictor.
\begin{equation}\label{eq:BYOL}
    \mathcal{L}_{BYOL}=-\frac{q(\hat{\matr{Y}_1})^\top \hat{\matr{Y}_2}}{||q(\hat{\matr{Y}_1})||\cdot||\hat{\matr{Y}_2}||}
\end{equation}

\noindent\textbf{Dense Contrastive Pretraining}:
BYOL is carried out at instance-level thus a global average pooling layer is appended at the end of backbone network to obtain a single feature vector for each image. For dense prediction tasks, e.g. segmentation, contrastive pretraining is carried out with spatial awareness. PixPro~\cite{xie2021propagate} proposed a dense contrative learning approach where global average pooling layer is removed and pixel-wise features are used for contrasting instead of image-wise features adopted in BYOL. 

\subsection{Target Pretraining for SSL}

\noindent\textbf{Contrastive Target Pretraining}:
In a regular transfer learning setting, we first assume a backbone network, parameterized by $\Phi$, is pretrained on a large dataset.
After pretraining, the model $\set{M}$ comes with parameter weights $\Phi_{pre}$. Then, $\Phi_{pre}$ is used to initialize the backbone network for the downstream/target task, e.g. finetuning on a smaller dataset or for semantic segmentation. As discussed, we believe directly re-using $\Phi_{pre}$ as initial weights for finetuning is sub-optimal. Instead, we propose to apply  constrative target pretraining on the target dataset. Specifically, we adopt BYOL contrastive loss with a regularization as below,
\begin{equation}
\resizebox{0.9\linewidth}{!}{
$
\begin{split}
    \min\limits_{\Phi}-\sum\limits_{\matr{X}_i\in\set{D}_L\cup\set{D}_U}\frac{q(\hat{\matr{Y}}_{i1})^\top \hat{\matr{Y}}_{i2}}{||q(\hat{\matr{Y}}_{i1})||\cdot||\hat{\matr{Y}}_{i2}||}
    + \lambda ||\Phi-\Phi_{pre}||^2_2
\end{split}
$
}
\end{equation}
We add a L2 weight regularization to avoid target pretrained weights to deviate too much from the initial weights. For a gradient-based optimization, the initial value for $\Phi$ is $\Phi_{pre}$. We find that by simply training enough iterations on target dataset, better weights can be obtained for semi-supervised finetuning. We denote the weights after $T$-epoch target pretraining as $\Phi_T$.

\noindent\textbf{Semi-Supervised Finetuning}
After target pretraining, the backbone weights $\Phi_T$ are now better for finetuning on target dataset. For simplicity, we denote the classifier as $\matr{Y}=h(\matr{Z};\Theta)$. Semi-supervised finetuning is then defined as optimizing the semi-supervised loss as,
\begin{equation}
\begin{split}
    \arg\min_{\Theta,\Phi}&\frac{1}{|\set{D}_L|}\sum_{\matr{X}_i,\matr{Y}_i\in\set{D}_L} \mathcal{L}_{CE}(h(f(\matr{X}_i;\Phi);\Theta),\matr{Y}_i) \\
    &+ \lambda\frac{1}{|\set{D}_U|}\sum_{\matr{X}_i\in\set{D}_U}\mathcal{L}_{con}(h(\matr{Z}_{i1};\Theta),h(\matr{Z}_{i2};\Theta))
    \end{split}
\end{equation}

The consistency loss $\mathcal{L}_{con}$ can be instantiated as off-the-shelf semi-supervised learners, e.g. FixMatch~\cite{sohn2020fixmatch}.



\subsection{Affine Augmentation for Segmentation}

One of the most critical ingredients in contrastive training is data augmentation. An investigation into effective data augmentation was carried out in \cite{chen2020simple} and many contrastive approaches follow the augmentation policies~\cite{grill2020bootstrap,xie2021propagate}. For a dense prediction task, \cite{xie2021propagate} developed a contrastive learning paradigm by comparing pixel-wise features and it turns out to be effective for detection and segmentation tasks. Therefore, we follow the dense contrastive learning strategy when target pretraining is applied to segmentation tasks. However, we believe the data augmentation strategies that work well for classification task may not be the best for segmentation.

CNNs are known to translation equivariant, therefore cropping and weak resizing as augmentation is not stronger enough to force the network learn useful information on unlabeled data. Under such weak geometric transformations, the network will simply output feature maps subject to the same transformation. Therefore, we propose to include a differentiable affine transformation~\cite{jaderberg2015spatial} to achieve more diverse geometric poses in augmented data. 

An affine transformation involves a combination of \textit{scaling}, \textit{rotation}, \textit{shearing} and \textit{translation}. Each time one affine transformation is constructed by randomly chosing each component with $50\%$ chance and drawing parameters from a predefined distribution. For input image, we apply the affine transformation after all non-differentiable data augmentation operations. As with STN~\cite{jaderberg2015spatial}, we use bilinear interpolation to estimate the pixel intensity after transformation. This transformation is differentiable w.r.t. input image and so does its inversed one.

We denote the transformation applied to image $\matr{X}$ as $t(\matr{X})$. To enable computing pixel-wise contrastive loss, we further apply the inverse transformation to the output of both online and target networks as $t^{-1}(f(t(\matr{X})))$. With the inverse transformation, predictions on two arbitrarily augmented images are directly comparable at pixel-level. 

\section{Experiments}
In this section, we introduce the datasets, experiment settings and then  provide benchmarking on semi-supervised finetuning with different initial weights. Finally, we provide hypothesis on the effectiveness of target pretraining from an adaptation perspective with empirical evidence.

\subsection{Datasets}
We demonstrate the effectiveness of proposed contrastive target pretraining strategy on four datasets. For classification tasks, we evaluate on CIFAR-10~\cite{krizhevsky2009learning}, CIFAR-100~\cite{krizhevsky2009learning} and SVHN~\cite{netzer2011reading}. We further evaluate on  additional semantic segmentation datasets, Pascal VOC 2012~\cite{pascal-voc-2012},  CityScapes~\cite{Cordts2016Cityscapes} and ISIC2017~\cite{xue2019robust}.
\textbf{CIFAR-10} is a image classification datasets with 50,000 training images and 10,000 testing images evenly categorized into 10 classes. For semi-supervised learning setting, we evaluate on 40, 250 and 1000 labeled data.
\textbf{CIFAR-100} extended from CIFAR-10 with 10 times more categories and the same amount of training/testing samples. Therefore, the categories are more fine-grained than CIFAR-10. For semi-supervised learning setting, we evaluate on 250 and 1000 labeled data.
\textbf{SVHN} is a street view house numbers recognition dataset with all numbers cropped and registered. There are 73257 digits for training, 26032 digits for testing respectively. For semi-supervised learning setting, we evaluate on 40 labeled data.
For CIFAR-10, CIFAR-100 and SVHN, we follow the standard data splits for semi-supervised evaluation as in~\cite{sohn2020fixmatch}.
\textbf{Pascal VOC 2012} has been widely adopted for semantic segmentation tasks. The dataset consists of 1464 training images  and 1449 validation images. Following the practice in \cite{french2019semi,xie2021propagate} we augment the training set with additional training images, resulting in 10582 training images in total. For semi-supervised learning, we evaluate on $1\%$, $2\%$ and $5\%$ of the labeled data.
\textbf{CityScapes} was collected from cars diving in urban environment with aim to segment objects commonly seen on the street. It has 2975 training and 500 validation samples, with 19 semantic classes. We evaluate on 100 and 372 labeled data.
\textbf{ISIC2017} is a medical image segmentation dataset detecting skin lesions. It has 2000 training images and features a significant difference from the source ImageNet dataset. We evaluate SSL at 10, 20 and 50 labeled data.

\subsection{Experiment Settings}
\noindent\textbf{Backbone Network}: We adopt Resnet18 and Resnet50~\cite{he2016deep} for classification and segmentation tasks respectively, as the weights pretrained on ImageNet is publicly available. 

\noindent\textbf{Data Augmentation}: Since backbone network is pretrained on ImageNet with input size $224\times224$, for target pretrain and finetuning, we first apply random cropping and then resize cropped region to $224\times224$. We further apply the same augmentations in BYOL~\cite{grill2020bootstrap} for target pretraining on classification tasks and with additional affine transformation for segmentation tasks.

\noindent\textbf{Contrastive Target Pretraining}: We adopt BYOL~\cite{grill2020bootstrap} and PixPro~\cite{xie2021propagate} for contrastive target pretraining for classification and segmentation respectively. We reduce the learning rate 10 times from the default ImageNet experiments in respective methods and further pretrain 300 epochs. For BYOL, we use the following hyperparameters, batchsize $256$, base learning rate $0.3$, momentum $0.9$ and total epochs $300$. For PixPro, we adopt the following hyperparameters, batchsize $8$, base learning rate $0.1$, momentum $0.9$ and total epochs $200$. The LARS optimizer is adopted for both pretraining. Weight regularization strength is fixed as $\lambda=1e-2$.

\noindent\textbf{Semi-Supervised Finetuning}:
In classification task we use FixMatch~\cite{sohn2020fixmatch} as the semi-supervised learner, we stick to the original hyperparameters for CIFAR-10, CIFAR-100 and SVHN reported in \cite{sohn2020fixmatch}. The following ones are adopted, the confidence threshold $\tau=0.95$, the unlabeled loss weight $\lambda_u=1$, the unlabeled to labeled data ratio in a minibatch $\mu=7$, the batchsize $B=64$ and the learning rate $1e-1$. SGD optimizer is adopted with momentum $\beta=0.9$.

For segmentation tasks, we adopt CutMix~\cite{french2019semi} as the semi-supervised leaner. We adopt the following hyperparameters, the initial learning rate $1e-5$, batchsize $20$/$8$ for PascalVOC and CityScapes respectively, iterations per epoch $1000$ and the number of total training epochs $40$. The Adam optimizer is adopted for both datasets. We randomly crop out a $321\times 321$ and $256\times 512$ region for training on PascalVOC and CityScapes respectively. For CityScapes, we downsample the image to $512\times512$ to allow more efficient training.

\subsection{Contrastive Target Pretraining}\label{sect:classificationtask}

\subsubsection{Classification Tasks.}
We first evaluate fully supervised baseline, Mean Teacher (MT)~\cite{tarvainen2017mean} and virtual adversarial training (VAT)~\cite{miyato2018virtual} as reference methods. We further evaluate two state-of-the-art SSL approaches, Self-Tuning~\cite{wang2021self} and FixMatch~\cite{sohn2020fixmatch} with different initial weights. As shown in Tab.~\ref{tab:classification}, we use ``Pretrain'' to refer to weights pretrained on large pretraining dataset, ``Trg. Pretr.'' indicates contrastive target pretraining and ``SSL'' indicates the semi-supervised learner. We make the following observations from this table. 

i) Target pretraining with BYOL is very effective regardless the initial weights being ImageNet supervised pretrained or from scratch. For example, if naive supervised finetuning is adopted, target pretraining improves $30\%$ and $15\%$ respectively from random initialized weights and ImageNet weights at CIFAR-10 with 40 labeled samples. The improvement is maintained with SelfTuning and FixMatch as semi-supervised learner on all datasets and labeling budgets. 

ii) As expected, the gap diminishes when more labeled data becomes available and SOTA semi-supervised learner is adopted for finetuning. For example, on CIFAR-10 with 250 and 1000 labeled samples with FixMatch finetuing, the improvement is negligible. This is due to more labeled data can mitigate the confirmation bias during semi-supervised learning. 

iii) Self-Tuning (ST)~\cite{wang2021self}, proposed to combine cross-entropy loss with a contrastive loss during finetuning.
ST was originally claimed to be superior to FixMatch at low-label regime~\cite{wang2021self}. However, through our extensive comparison, we conclude that the advantage of ST is mainly due to a better initial weights. For example, with the best initial weights (IN Sup+BYOL), ST is still far behind FixMatch on all datasets. This implies that developing more effective pseudo-labeling algorithm for SSL is still one of the most effective way to exploit unlabeled data. Nevertheless, a good initial weights are equally important, combining finetuning with semi-supervised learning should receive more attention for more data-efficient SSL. 


\begin{table*}[htbp]
  \centering
  \caption{Evaluation of the impact of further contrastive pretraining by semi-supervised finetuning on classification tasks. IN Sup indicates supervise pretrained on ImageNet. The number in parenthesis ($\pm$X) is the improvement brought by target pretraining. 
  }
  \resizebox{1\linewidth}{!}{
    \begin{tabular}{lllllllll}
    \toprule
          &       &       & \multicolumn{3}{c}{CIFAR-10} & \multicolumn{2}{c}{CIFAR-100} & SVHN \\
\cmidrule(lr){4-6} \cmidrule(lr){7-8} \cmidrule(lr){9-9}  Pretrain & Trg. Pre. & SSL & \multicolumn{1}{l}{\#40} & \multicolumn{1}{l}{\#250} & \multicolumn{1}{l}{\#1000} & \multicolumn{1}{l}{\#400} & \multicolumn{1}{l}{\#2500} & \multicolumn{1}{l}{\#40} \\
    \midrule
    -     & -     & MT~\cite{tarvainen2017mean}   & 27.34 & 47.29 & 65.48 & 10.36 & 37.49 & 27.73 \\
    -     & -     & VAT~\cite{miyato2018virtual}   & 31.81 & 57.02 & 75.61 & 12.85 & 32.32 & 15.21 \\
    \midrule
    -     & -     & -     & 18.41 & 29.76 & 44.96 & 8.67  & 21.69 & 10.47 \\
    -     & BYOL  & -     & 49.44~(+31.03) & 66.33~(+36.57) & 75.15~(+30.19) & 15.18~(+6.51) & 35.20~(+13.51) & 18.25~(+7.78) \\
    IN Sup & -     & -     & 49.69 & 77.38 & 85.56 & 30.85 & 57.36 & 16.92 \\
    IN Sup & BYOL  & -     & 65.13~(+15.44) & 82.07~(+4.69) & 87.64~(+2.08) & 34.63~(+3.78) & 58.65~(1.29) & 35.82~(+18.90) \\
    \midrule
    -     & -     & SelfTuning~\cite{wang2021self} &   27.89    &   45.84    &  61.06     & 9.17  & 23.72 &  13.49 \\
    -     & BYOL  & SelfTuning &    49.76~(+21.87)   &   73.70~(+27.86)    &    82.11~(+21.05)   & 19.05~(+9.88) & 41.40~(+17.68) & 22.23~(+8.74) \\
    IN Sup & -     & SelfTuning & 48.83 & 82.56 & 89.86 & 35.56 & 62.86 & 45.84 \\
    IN Sup & BYOL  & SelfTuning & 62.57~(+13.74) & 84.71~(+2.15) & 90.83~(+0.97) & {38.36}~(+2.80) & {67.31}~(4.45) & 46.95~(+1.11) \\
    \midrule
    -     & -     & FixMatch~\cite{sohn2020fixmatch} & 60.75 & 91.32 & 93.01 & 29.85 & 59.12 & 32.76 \\
    -     & BYOL  & FixMatch & 75.51~(+14.76) & 91.55~(+0.23) & 94.19~(+1.18) & 40.14~(+10.29) & 63.34~(+4.22) & 36.94~(+4.18) \\
    IN Sup & -     & FixMatch & 75.28 & 94.22 & 95.82 & 44.34 & 71.81 & 79.89 \\
    IN Sup & BYOL  & FixMatch & {90.06}~(+14.78) & {94.99}~(+0.77) & {95.48}~(-0.34) & {46.23}~(+1.89) & {72.19}~(+0.38) & {92.97}~(+13.08) \\
    \bottomrule
    \end{tabular}%
    }
  \label{tab:classification}%
\end{table*}%

\subsubsection{Segmentation Tasks.} We further present semi-supervised finetuning results on semantic segmentation task in Tab.~\ref{tab:segmentation}. We compare three state-of-the-art semi-supervised segmentation approaches, CutMix~\cite{french2019semi}, ClassMix~\cite{olsson2021classmix} and CLMB~\cite{alonso2021semi}. We make the following observations from the segmentation experiments. 

i) Contrastive target pretraining demonstrates consistent effectiveness on semantic segmentation task. The improvement is particularly large for CutMix at low-label regime, e.g. Pascal VOC $1\%$ and CityScapes \#100 labeled data. This again implies the bottleneck of from-scratch semi-supervised finetuning is prone to confirmation bias when labeled data is small. As a result, better initial weights is required  

ii) It is also surprising to see with the target pretraining step, the final results of CutMix can sometimes match or slightly outperform its counterpart with ImageNet pretrained weights. It is worth noting that no labels are required for PixPro to pretrain ImageNet, thus a substantial amount of labeling efforts is saved under this setting.

\begin{table*}[htbp]
  \centering
  \caption{Evaluation of Semi-supervised finetuning with different model initial weights. IN PixPro indicates weights pretrained with PixPro on ImageNet.}
  \resizebox{0.6\linewidth}{!}{
    \begin{tabular}{lllccccc}
    \toprule
          &       &       & \multicolumn{3}{c}{PascalVOC} & \multicolumn{2}{c}{CityScapes} \\
\cmidrule(lr){4-6} \cmidrule(lr){7-8}    Pretrain & Trg.Pre. & SSL   & 1\%   & 2\%   & 5\%   & \#100 & \#372 \\
    \midrule
    IN Sup   & -     & CutMix~\cite{french2019semi} & 53.79 & 64.81 & 66.48 & 57.89 & 65.64 \\
    IN Sup   & -     & ClassMix~\cite{olsson2021classmix} & 40.64 & 52.84 & 59.82 & -     & - \\
    IN Sup   & -     & CLMB~\cite{alonso2021semi}  & -     & 63.40 & 69.10 & 64.90 & 70.00 \\
    \midrule
    IN PixPro & -     & -     & 38.20 & 49.95 & 60.35 & 52.99 & 63.05 \\
    IN PixPro & PixPro & -     & 41.35 & 52.97 & 63.11 & 53.26 & 63.26 \\
    IN PixPro & -     & CutMix & 49.58 & 61.63 & 67.61 & 58.30 & 68.21 \\
    IN PixPro & PixPro & CutMix & 52.20 & 61.25 & 68.59 & 59.20 & 68.02 \\
    \bottomrule
    \end{tabular}%
    }
  \label{tab:segmentation}%
\end{table*}%

\subsubsection{Transferring to Medical Image Segmentation}
We further evaluate target pretraining on target datasets with a large domain gap by performing an experiment {transferring a model pretrained on ImageNet to a skin lesion segmentation dataset, ISIC2017~\cite{codella2018skin}}. We follow the settings proposed in CutMix and consider labeling budgets of 10, 20 and 50 labeled samples. We use 1000 epochs of target pretraining implemented on all 2000 training examples of ISIC2017. Results are shown in Tab.~\ref{tab:ISIC2017} -- the consistent improvement in the low-label regime further validates the effectiveness of target pretraining.

\begin{table}[htbp]
  \centering
  \caption{Target pretraining on skin lesion segmentation dataset (ISIC2017). Result in () was reported in~\cite{french2019semi}.}
  \vspace{-0.3cm}
  \resizebox{0.7\linewidth}{!}{
    \begin{tabular}{cccc}
    \toprule
    \#Labeled & Pretrain & Trg. Pre. & mIoU \\
    \midrule
    \multirow{2}[2]{*}{10} & IN Sup    & -     & 73.80 \\
          & IN Sup    & BYOL  & 77.74 \\
    \midrule
    \multirow{2}[2]{*}{20} & IN Sup    & -     & 82.93 \\
          & IN Sup    & BYOL  & 82.96 \\
    \midrule
    \multirow{2}[2]{*}{50} & IN Sup    & -     & 85.99 (74.57) \\
          & IN Sup    & BYOL  & 86.02 \\
    \bottomrule
    \end{tabular}%
    }
  \label{tab:ISIC2017}%
\end{table}%

\begin{figure}[!h]
    \centering
    \subfloat[]{\includegraphics[width=0.53\linewidth]{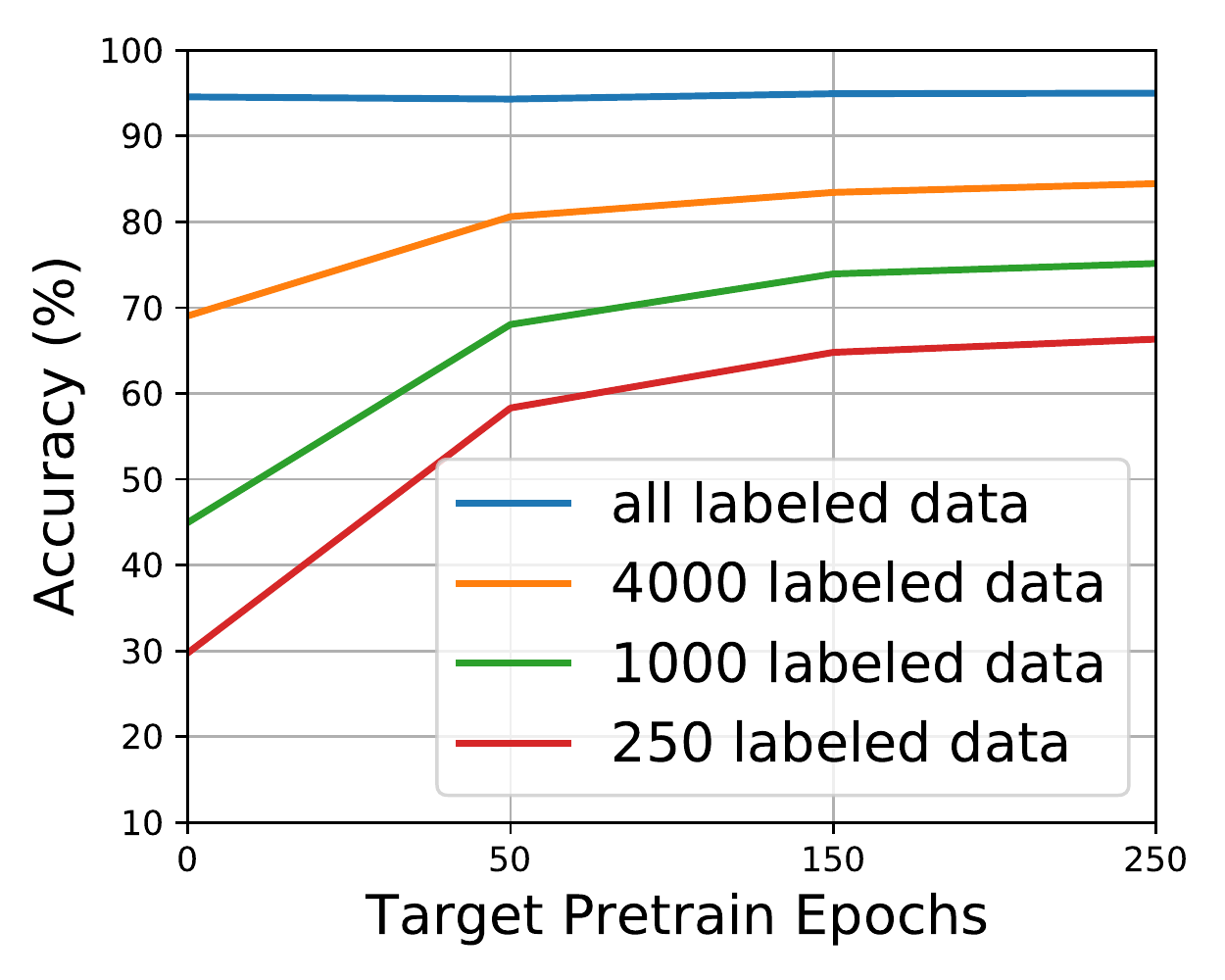}}
    \subfloat[]{\includegraphics[width=0.53\linewidth]{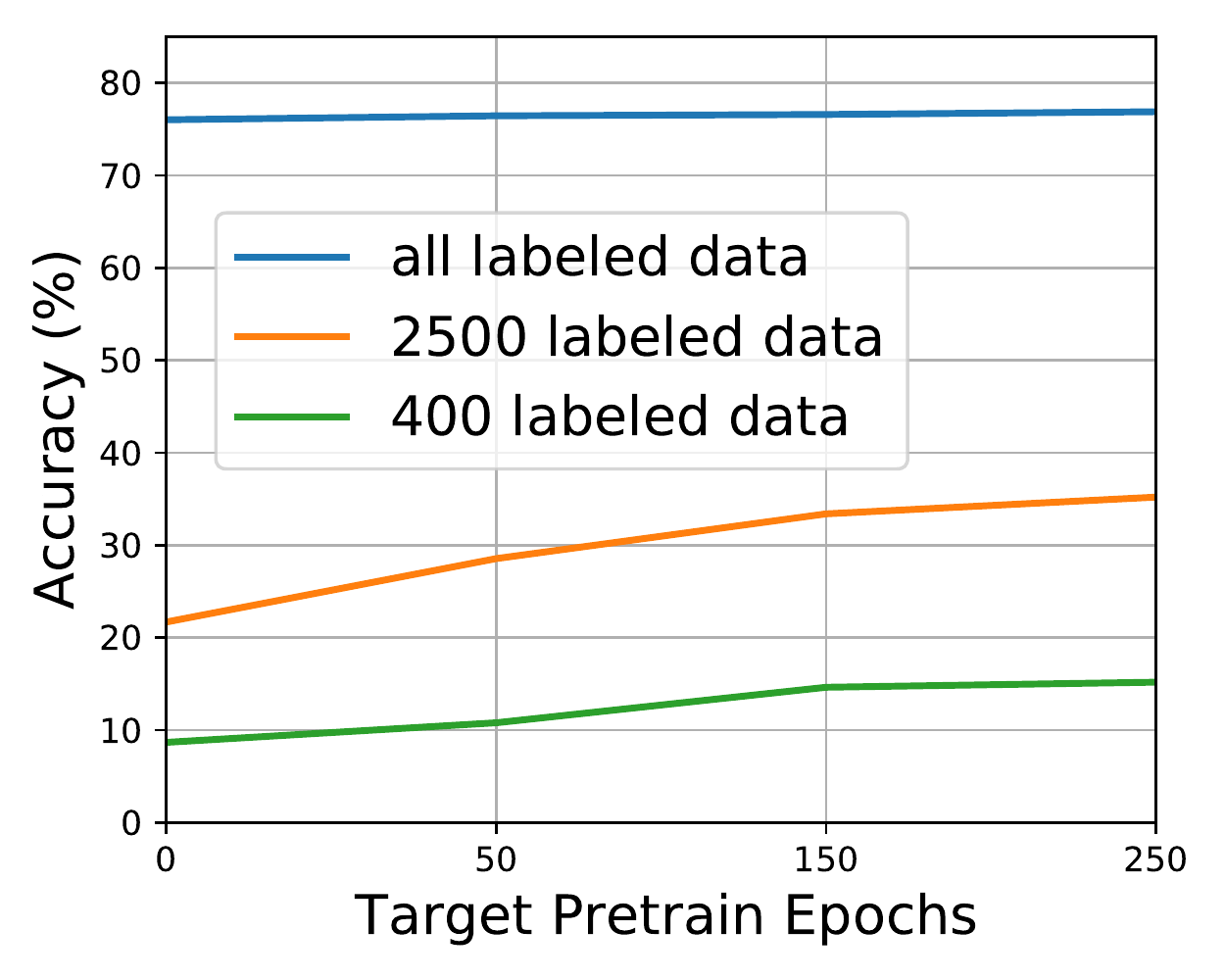}}
    \caption{The impact of target pretraining on CIFAR-10 and CIFAR-100 datasets.}
    \label{fig:StepsFurtherPretrain}
\end{figure}

\subsection{Ablation Study}
In this section, we carry out ablation study to verify the importance of weight regularization and affine transformation as augmentation. We carry out ablation on PascalVOC segmentation task. As shown in Tab.~\ref{tab:ablation}, we observe both weight regularization and affine transformation as augmentation improve the generalization of further pretrained model weights.

\begin{table}[htbp]
  \centering
  \caption{Ablation study of weight regularization and differentiable affine transformation as augmentation.}
  \resizebox{0.85\linewidth}{!}{
    \begin{tabular}{ccccc}
    \toprule
          &       & \multicolumn{3}{c}{PascalVOC} \\
\cmidrule{3-5}    \multicolumn{1}{l}{Weight Reg.} & \multicolumn{1}{l}{Affine Aug} & 1\%   & 2\%   & 5\% \\
    \midrule
    -     & -     & 38.20 & 49.95 & 60.35 \\
    \checkmark     & -     & 40.32 & 52.00 & 62.39 \\
    \checkmark     & \checkmark     & 41.35 & 52.97 & 63.11 \\
    \bottomrule
    \end{tabular}%
    }
  \label{tab:ablation}%
\end{table}%

\begin{figure*}[!htb]
    \centering
    \includegraphics[width=1\linewidth]{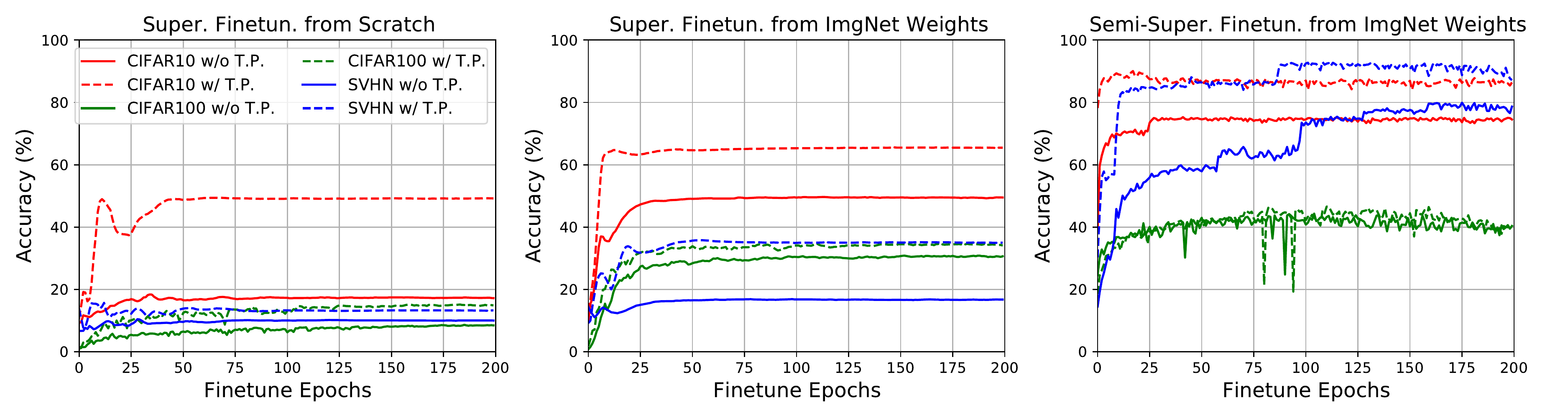}
    \caption{Comparing finetuning with different initial weights, ``w/o T.P.'' and ``w/ T.P.'' indicate with and without target pretraining respectively. Left and middle plots illustrate supervised finetuning from weights with target pretraining from random weights and ImageNet  weights respectively. Right plot illustrates semi-supervised finetuning.}
    \label{fig:convergence}
\end{figure*}

We perform additional evaluations on alternative contrastive learning and semi-supervised learning methods on CIFAR-10 with 40 labeled samples. We perform target pretraining with MoCo~v2~\cite{chen2020improved} in addition to BYOL on CIFAR-10 with ImageNet pretrained weights and SSL finetuning with UDA~\cite{ganin2015unsupervised} and MeanTeacher in addition to FixMatch. From Tab.~\ref{tab:additional}, we observe consistent improvements with alternative contrastive learning methods and we observe the advantage of target pretraining with additional SSL methods as well.

\begin{table}[htbp]
  \centering
  \caption{Additional contrastive learning and semi-supervised finetuning methods.}
    \begin{tabular}{cccccc}
    \toprule
          &       & \multicolumn{4}{c}{Semi-Supervised Finetune} \\
    \cmidrule(lr){3-6}
    Pretrain & Trg. Pre. & -     & FixMatch~\cite{sohn2020fixmatch} & UDA~\cite{ganin2015unsupervised}   & MT~\cite{tarvainen2017mean} \\
    \midrule

    IN Sup    & -     & 49.69 & 75.28 & 62.94 & 48.98 \\
    IN Sup    & BYOL~\cite{grill2020bootstrap}  & 65.13 & 90.06 & 78.43 & 61.55 \\
    IN Sup   & MoCo~v2~\cite{chen2020improved} & 71.13 & 90.37 & 87.04 & 69.02 \\
    \bottomrule
    \end{tabular}%
  \label{tab:additional}%
\end{table}%

\subsection{Further Insights into Target Pretraining}
We discuss in this part the hypotheses of why this simple target pretraining is so effective. Some early works managed to explain deep neural network pretraining from providing a regularization effect on model parameters~\cite{erhan2010does}. For example, pretraining will initialize the parameters in the basin of attraction of a good local optimum. Therefore, the following supervised finetuning will be easier to generalize. Another recent empirical study~\cite{he2019rethinking} argued that when there is enough labeled data, training from scratch is comparable to finetuning from pretrained weights. However, a breakdown point was observed in the low labelling regime, indicating good initial weights is necessary when labeled data is low.

In contrast to the previous attempts to explain, we would like to analyze from a covariate shift point of view. W.l.o.g, we describe a general supervised pretraining and finetuning example. The conclusion also applies to contrastive pretrain and semi-supervised finetuning.
For simplicity, we denote an additional unlabelled pretraining dataset as $\set{D}_{pre}$. The objective of supervised pretraining can be seen as minimizing the following negative log likelihood, where we can decompose the conditional probability into the backbone $p(\matr{Z}|\matr{X};\Phi)$ and classifier $p(\matr{Y}|\matr{Z};\Theta)$.
\begin{equation}
\resizebox{0.9\linewidth}{!}{
$
    \min_{\Theta_{pre},\Phi_{pre}} \mathbb{E}_{\matr{X}\sim p_{pre}} -\log p({\matr{Y}}|\matr{Z};\Theta_{pre})p(\matr{Z}|\matr{X};\Phi_{pre})\\
$
}
\end{equation}

The objective of finetuning can be written as below.
\begin{equation}
\begin{split}
    &\min_{\Theta_{ft},\Phi_{ft}} \mathbb{E}_{\matr{X}\sim p_{ft}} -\log p(\matr{Y}|\matr{Z};\Theta_{ft})p(\matr{Z}|\matr{X};\Phi_{ft})\\
\end{split}
\end{equation}

The data distribution between pretraining and finetuning stages are often not identical thus subject to a covariate shift, i.e. $p_{pre}(\matr{X})\neq p_{ft}(\matr{X})$. It is reasonable to believe the optimal backbone networks subject to pretraining and finetuning objectives respectively are not identical $p(\matr{Z}|\matr{X};\Phi_{ft}^*) \neq p(\matr{Z}|\matr{X};\Phi_{pre}^*)$ either.
As a result, directly re-using $p(\matr{Z}|\matr{X};\Phi_{pre})$ for the supervised finetuning task would be sub-optimal. Further pretrain the backbone on the whole target dataset and the backbone parameters are initialized by the pretrained parameters, i.e. $\Phi_{ft}^0=\Phi_{pre}^*$, as below.
\begin{equation}
    \min_{\Phi_{ft}} \sum_{\matr{X}\in \set{D}_L\cup\set{D}_U} -\log p({\matr{Y}}|\matr{Z};\Theta_{pre})p(\matr{Z}|\matr{X};\Phi_{ft})\\
\end{equation}\label{eq:further}

Compared with direct finetuning from $\Phi_{pre}$, target pretrained weights $\Phi_{ft}$ is first updated on the whole target dataset $\set{D}_L\cup\set{D}_U$. With covariate shift being eliminated, $\Phi_{ft}$ will move towards the optimal weights for target task. Eventually, $\Phi^*_{ft}$ as initial weights for supervised finetuning can ease the
difficulty in optimization for finetuning. 
To validate the above hypotheses we carry out the following empirical studies.

\subsubsection{Target Pretraining as Adaptation}\label{sect:TransferWeights}
In this experiment, we validate the hypothesis of target pretraining as an adaptation of weights pretrained on generic dataset to specific target dataset. We keep all the target pretraining protocols the same with Sect.~\ref{sect:classificationtask}, an ImageNet pretrained weights are taken as initial weights, target pretraining is carried out on three classification datasets respectively.
A cross-dataset finetuning experiment is carried out by transferring the target pretrained weights across CIFAR-10, CIFAR-100 and SVHN, with results presented in Tab.~\ref{tab:transferweights}. We make the following observations. First, transferring weights target pretrained on CIFAR-10/100 to finetuning on SVHN yields moderate improvement from without target pretraining ($16.92\%\rightarrow19.71\%/20.06\%$), while they are still behind target pretrained on SVHN itself. Due to the large distribution gap between CIFAR-10/100 (natural object images) and SVHN (digit images), this result suggests target  pretraining is essentially adapting weights to target data distribution. Moreover, transferring SVHN target pretrained weights to CIFAR10/100 produces much inferior results than even without target pretraining and transferring between CIFAR-10 and CIFAR-100 gives a relatively smaller drop from within dataset pretraining. Both of the above two observations indicate target pretraining must be carried out on the distribution
of target dataset which again validates the adaptation hypothesis.

\begin{table}[htbp]
  \centering
  \caption{Transferring target pretrained weights across datasets. Accuracy (\%) is reported.} 
  \resizebox{0.9\linewidth}{!}{
        \begin{tabular}{llccc}
    \toprule
          &       & \multicolumn{3}{c}{Finetune } \\
\cmidrule{3-5}    Pretrain & Trg. Pre. & CIFAR-10 & CIFAR-100 & SVHN \\
    \midrule
    IN Sup & -     & 49.69 & 30.85 & 16.92 \\
    IN Sup & CIFAR-10 & 65.13 & 29.51 & 19.71 \\
    IN Sup & CIFAR-100 & 49.46 & 34.63 & 20.06 \\
    IN Sup & SVHN  & 24.91 & 12.12  & 35.82 \\
    \bottomrule
    \end{tabular}%
    }
  \label{tab:transferweights}%
\end{table}%

\subsubsection{Target Pretraining Improves Class Discovery}
We further investigate how target pretraining helps adapt to better initial weights by evaluating unsupervised class discovery (clustering). As an alternative approach towards unsupervised feature learning, clustering has been employed to generate pseudo-labels for representation learning~\cite{caron2018deep,caron2020unsupervised}. Therefore, if better clustering results are identified it means the features are potentially more useful for finetuning. 
Specifically, we first denote the cluster indices as $\matr{Y}_{c}\in\{0,1\}^{N\times K}$ and the ground-truth classification label as $\matr{Y}\in\{0,1\}^{N\times K}$, where $N$ is the number is samples and $K$ is the number of classes/clusters. For both $\matr{Y}_c$ and $\matr{Y}$, there is only a single element with 1 at each row. Since clustering is invariant to the permutation of cluster index the assignment of cluster index to class label is unknown. To calculate clustering accuracy, we first find the following permutation of cluster indices,
\begin{equation}
\begin{split}
    &\matr{P}\in\{0,1\}^{K\times K}\\ 
    s.t.\quad \forall i,j\in\{&0,\cdots,K\},\quad \sum\vect{p}_i=1,\;\sum\vect{p}_j=1
\end{split}
\end{equation}

The following problem is formulated to find the best permutation that maximizes the accuracy, where $\circ$ is an elementwise product. 

\begin{equation}
\begin{split}
    &\max_{\matr{P}} \frac{1}{N} ||vec(\matr{Y}_{c}\matr{P} \circ \matr{Y})||_1\\
    \rightarrow &\max_{\matr{P}} \frac{1}{N} vec(\matr{Y}_{c}\matr{P})^\top vec(\matr{Y})\\
    \rightarrow &\max_{\matr{P}} \frac{1}{N} tr(\matr{P}^\top\matr{Y}_c^\top\matr{Y})
\end{split}
\end{equation}

The above problem can be solved by Hungarian algorithm where the cost matrix is $-\matr{Y}_c^\top\matr{Y}$.
We eventually report the performance as the classification described above. 
With results presented in Tab.~\ref{tab:clustering}, we observe consistently better clustering results for target pretrained features suggesting the effectiveness of target pretraining as good initial weights for representation learning.

\begin{table}[htbp]
  \centering
  \caption{Evaluation of target pretraining by clustering accuracy (\%). Results on training and testing data are delimited by /. }
    \resizebox{1\linewidth}{!}{
    \begin{tabular}{llccc}
    \toprule
    Pretrain & Trg. Pretr. & Cifar-10 & Cifar-100 & SVHN \\
    \midrule
    IN Sup & -     & 46.31/45.66 & 6.40/6.74 & 12.96/12.64 \\
    IN Sup & BYOL  & 47.17/48.02 & 7.91/7.44 & 15.96/16.23 \\
    \bottomrule
    \end{tabular}%
    }
  \label{tab:clustering}%
\end{table}%

\subsubsection{TSNE Visualization.}
\begin{figure}
    \centering
    \includegraphics[width=1.04\linewidth]{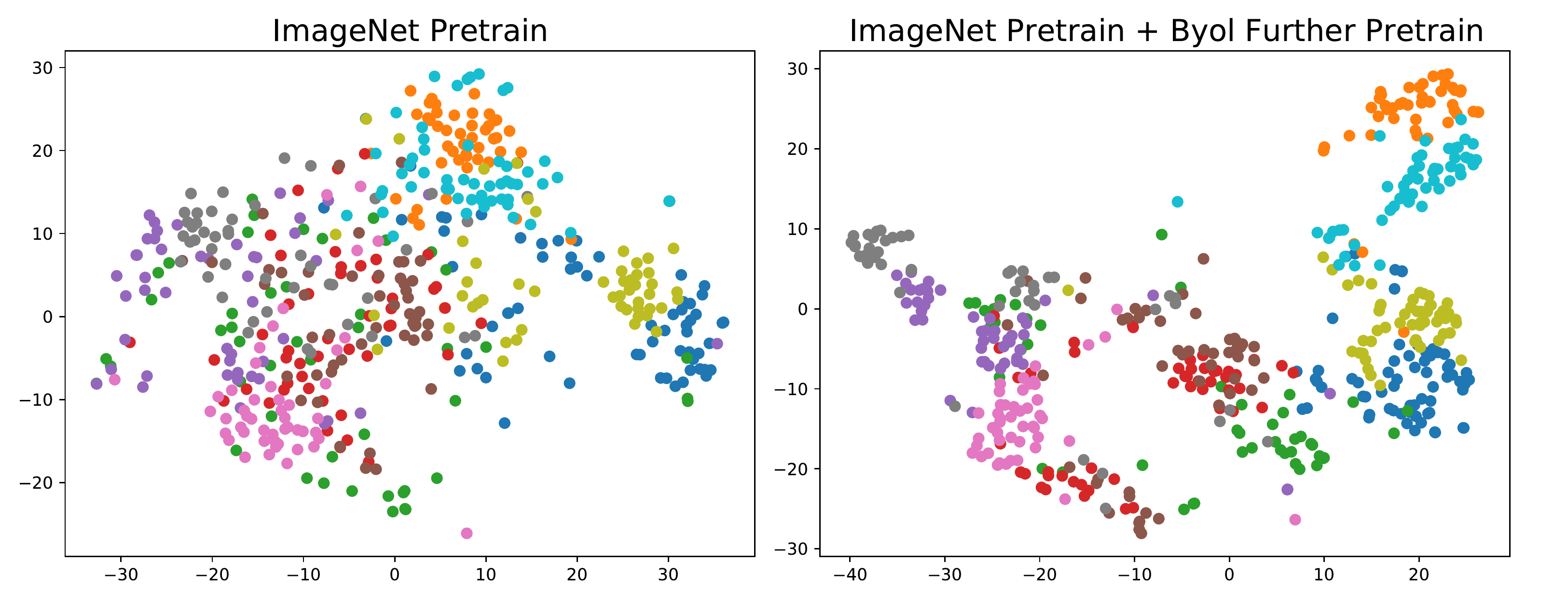}
    \caption{TSNE visualization of 500 randomly selected test samples from CIFAR-10 before (left) / after (right) target pretraining.}
    \label{fig:tsne}
\end{figure}

We further visualize feature representations after target pretraining on target dataset to draw qualitative insight into its effect. Specifically, we collect 500 randomly selected testing samples from CIFAR-10 with features as the output of last layer in backbone after global average pooling. We compared two set of features, the first is collected from ResNet18 with ImageNet pretrained weights and another is from ImageNet pretrained weights plus target pretraining. We concatenate both sets of features and project them into 2D via TSNE~\cite{van2008visualizing} for visualization. As shown in Fig.~\ref{fig:tsne}, we see a noticeable clustering structure after target pretraining on target data, demonstrating the effectiveness of target pretraining.

\subsubsection{Impact of Target Pretraining Steps.}
We study the impact of target pretraining w.r.t. the number of labeled data and steps of target pretraining. Specifically, we evaluate supervised finetuning performance on CIFAR-10 with 250, 1000, 4000 and all labeled data. The initial weights for finetuning is evaluated from 0 to 250 epochs. The results are illustrated in Fig.~\ref{fig:StepsFurtherPretrain}. First, target pretraining is effective after at least 200 epochs training, suggesting enough training iterations is required for a good adaptation to target dataset. Second, the improvement brought by target pretraining is consistently better at lower-labeling regime. This aligns with many previous investigations into the impact of pretraining~\cite{he2019rethinking}.

\subsubsection{Comparing Convergence}
As discussed, having good initialization allows faster finetuning with better results after convergence. In this section, we compare the convergence between with and without contrastive target pretraining on CIFAR-10 \#40 labeled data, CIFAR-100 \#400 labeled data and SVHN \#40 labeled data. Specifically, we evaluate both supervised finetuning and semi-supervised finetuning with different initial weights and visualize the results in Fig.~\ref{fig:convergence}. In Fig.~\ref{fig:convergence}~(left), we first show the test accuracy curves on three classification datasets over 200 training epochs. Initial weights are randomly initialized before target pretraining. For all three datasets, it is very clear that with target pretraining (dashed line), the converged accuracy is substantially higher. In Fig.~\ref{fig:convergence}~(middle), with ImageNet supervised pretrained weights, target pretraining (dashed line) demonstrates superior performance with significant margin on CIFAR-10 and SVHN. Finally, in Fig.~\ref{fig:convergence}~(right), we compare the convergence of FixMatch semi-supervised finetuning with different initial weights. There is a very clear margin for CIFAR-10 and SVHN when target pretraining is included while the margin is smaller for CIFAR-100 probably because more labeled data (\#400 labeled) is available in CIFAR-100 experiment. We also notice that all methods have converged within 200 epochs, thus making this comparison fair. All these observations again validate that good initial weights play an important role in SSL finetuning at low-label regime and target pretraining is an effective way to produce better initial weights.



\section{Conclusion}
Motivated by the recent studies into combining transfer learning with semi-supervised learning, we first revealed that a good initial weights are actually accountable for the substantial improvement on semi-supervised finetuning. We demonstrate that with the same initial weights, semi-supervised learning based on pseudo-labeling is still a better option. We further discover that due to the covariate shift between pretrain and target datasets, direct finetuning from weights pretrained on separate large dataset is not optimal. A contrastive target pretraining step is proposed to adapt the weights to target dataset. We demonstrate on multiple classification and segmentation datasets that semi-supervised finetuning can benefit substantially from the adapted weights at low-label regime. This study encourages people to rethink the relation between transfer learning and semi-supervised learning and contrastive learning could be a good way to synergistically combine both.


\bibliographystyle{IEEEtran}
\bibliography{reference}

\vfill

\end{document}